\documentclass[journal]{IEEEtran}

\usepackage{cite}
\usepackage{multirow}

\ifCLASSINFOpdf
\usepackage[pdftex]{graphicx}
\else
\usepackage[dvips]{graphicx}
\fi

\usepackage{subfig}
\usepackage{amsmath}
\usepackage{units}
\usepackage[hyphens]{url}

\usepackage[colorinlistoftodos]{todonotes}

\begin{document}

\title{Playing Against the Board: Rolling Horizon Evolutionary Algorithms Against Pandemic}
\author{Konstantinos~Sfikas and Antonios~Liapis,~\IEEEmembership{Member,~IEEE}
\thanks{All authors are with the Institute of Digital Games, University of Malta, Msida, MSD 2080, Malta}
}

\markboth{IEEE Transactions on Games}%
{Playing Against the Board: How Artificial Intelligence Can Beat Pandemic (the Board Game)}

\maketitle

\begin{abstract}
Competitive board games have provided a rich and diverse testbed for artificial intelligence. This paper contends that collaborative board games pose a different challenge to artificial intelligence as it must balance short-term risk mitigation with long-term winning strategies. Collaborative board games task all players to coordinate their different powers or pool their resources to overcome an escalating challenge posed by the board and a stochastic ruleset. This paper focuses on the exemplary collaborative board game Pandemic and presents a rolling horizon evolutionary algorithm designed specifically for this game. The complex way in which the Pandemic game state changes in a stochastic but predictable way required a number of specially designed forward models, macro-action representations for decision-making, and repair functions for the genetic operations of the evolutionary algorithm. Variants of the algorithm which explore optimistic versus pessimistic game state evaluations, different mutation rates and event horizons are compared against a baseline hierarchical policy agent. Results show that an evolutionary approach via short-horizon rollouts can better account for the future dangers that the board may introduce, and guard against them. Results highlight the types of challenges that collaborative board games pose to artificial intelligence, especially for handling multi-player collaboration interactions.
\end{abstract}


\IEEEpeerreviewmaketitle

\section{Introduction}\label{sec:introduction}

\IEEEPARstart{P}{erhaps} some of the most iconic and publicly resonant moments for Artificial Intelligence have been competitions with human masters in board game such as Chess \cite{Newborn1997KasparovVD} and Go \cite{Gershgorn2016Alphago}. Modern-day academic research has explored a diverse set of algorithms for playing board games \cite{guhe2014catan,silva2017ticket,mazyad2015jack} and card games \cite{Cowling2012EnsembleDI,Sephton2014HeuristicMP} which focus on player-versus-player competition or, at best, team-versus-team competition\cite{WaltonRivers2017EvaluatingAM,Summerville2019CodenamesAI}. However, a large portion of modern board games eschew competition between players and instead invite \emph{collaborative} play, where all players must work together to survive (and win) against a rule-based system which presents an escalating challenge. Common design patterns for such collaborative games are (a) player roles specializing in certain tasks, (b) a rule-based system with high stochasticity (via drawn cards or dice) that introduces more and more complications and challenges to the game state, (c) a race against time for players to achieve victory, and (d) a dilemma between performing actions that mitigate current threats and actions that lead to victory. \emph{Pandemic} (Z-Man, 2008) is among the most popular collaborative board games and is fairly straightforward to play: players take different specialized roles and strive to cure diseases while more and more cities on the board are infected with disease cubes. Players must balance between removing disease cubes (to stop the game from ending) while also collecting cards in order to cure diseases (to win the game). What makes {Pandemic} particularly interesting is that infections are not chosen completely randomly; {Pandemic} implements a clever system of recycling past infected cities. This means that players can anticipate the next few cities that will be infected (but not the order in which they will be infected) and strategize how best to minimize the risk.

This paper argues that collaborative board game play poses its own set of challenges to Artificial Intelligence. While competitive play challenges AI to anticipate what the other player might do or how best to block another player, collaborative board game play challenges AI to best coordinate with other players. When all players are controlled by AI, a plan can be formulated for every player (by a single controller) and executed to the letter. This is actually how human players also handle a collaborative board game by making a strategy for every player's move and executing it (or replan, if circumstances change). The AI challenge of collaborative board game play is thus not to align each player's goal (as a single controller can control every player) but instead (a) to balance between short-term damage control and long-term strategies that win the game, (b) to optimally take advantage of different players' roles and special abilities, and (c) to anticipate the best- and worst-case scenarios of upcoming events and how they will affect the game state. Due to the stochastic nature of escalating threats posed by the game system, human players similarly perform risk assessment and mitigation in the hopes of avoiding the worst outcomes which usually lead to a loss.

This paper expands on recent experiments in applying a Rolling Horizon Evolutionary Algorithm (RHEA) for playing Pandemic \cite{sfikas2020collaborative} which primarily explored the impact of different optimistic and pessimistic functions for evaluating the game-state. This paper expands this work by exploring the impact of different parameters such as mutation operators, generations, and plan horizon to the algorithm's performance. A broader analysis of the robustness of the algorithm in random game setups at different difficulty levels also highlight the strengths and limitations of this approach. 
The structure of the paper is as follows: Section \ref{sec:relatedwork} discusses related work on board game play and RHEA methods; Section \ref{sec:game} explains the components, rules, and actions available in Pandemic; Section \ref{sec:methodology} presents the game state, action representation and the RHEA algorithm; Section \ref{sec:experiment} presents a broad set of experiments assessing the algorithm's performance under different circumstances; Section \ref{sec:discussion} discusses the findings and suggests future work while Section \ref{sec:conclusion} concludes the paper.
The codebase for the Pandemic testbed and agents is publicly available at: \url{https://github.com/konsfik/Pandemic-AI-Framework}.

\section{Related Work}\label{sec:relatedwork}

Until the early 2000s, game playing AI was mostly applied to deterministic, fully observable, adversarial games such as chess, hex and Othello. Early board game play focused on \emph{MiniMax} and its variants, such as $\alpha-\beta$ pruning \cite{Knuth1975AnAO}. IBM's Deep Blue applied \emph{MiniMax} to defeat the chess champion Garry Kasparov in 1997 \cite{Campbell2002DeepB}. In the early 2000s, Monte Carlo Tree Search (MCTS) started to appear as an alternative to \emph{MiniMax}. MCTS operates via an iterative process of selection, expansion, simulation and backpropagation that gradually expands the game tree in a stochastic manner \cite{Browne2012ASO}. MCTS enabled several boardgame-playing programs \cite{arneson2010hex,Gelly2012TheGC,zhuang2015dotsandboxes,tang2016gomoku,yang2016connect6} that could perform comparably to professional human players, culminating in the success of DeepMind's AlphaGo \cite{DavidSilver2016Mtgo} against multiple champions in the board game Go.

The Rolling Horizon Evolutionary Algorithm \cite{perez2013rhea} (RHEA) is an evolutionary planning algorithm which applies evolution directly onto the decision-making process, similar to how MCTS uses roll-outs and the generative model. RHEA operates as follows: starting from a specific game state, RHEA evolves a set of action sequences, using the game's forward model to estimate their future consequences. When evolution ends, RHEA selects the first action of the best sequence and applies it to the game state. This process is performed repeatedly until the game is over. 
RHEA has been applied to a number of digital games \cite{perez2016gvgai,gaina2017enhancements,Resnick2018PommermanAM,Liebana2019AnalysisOS}, as well as modern board games such as \emph{Splendor} (Space Cowboys, 2014) \cite{Bravi2019splendor}. 
In many cases, the performance of RHEA is at least comparable to that of MCTS. RHEA seems to operate better in coarse-grained state representations, e.g. when using macro actions instead of single actions \cite{perez2013rhea}. RHEA is dependent on a fitness function, which may seem like a drawback compared to MCTS which---at least in theory---can function without any domain knowledge \cite{Browne2012ASO}. However, machine learning \cite{Tong2019EnhancingRH} or Monte-Carlo rollouts \cite{gaina2017enhancements} can replace biases of an ad-hoc fitness function. Another control point for the algorithm's optimization is the population-seeding policy, which can provide a better starting point in the evolutionary process. For instance, Gaina \emph{et al.} \cite{gaina2017seeding} seed a population using a 1-step-lookahead method or a MCTS method and show that both shortcuts have a positive effect on the algorithm's performance, especially when the computational resources are limited. Other seeding options include re-using populations from previous decision points \cite{gaina2017enhancements,santos2018improved} or employing a machine-learned policy \cite{Tong2019EnhancingRH}.

As noted in Section \ref{sec:introduction}, modern commercial board games have been a rich and diverse testbed for modern AI methods. MCTS has been applied to numerous board games such as \emph{Settlers of Catan} (Kosmos, 1995) \cite{szita2009catan,guhe2014catan}, \emph{Thurn and Taxis} (Hans im Gl\"{u}ck, 2006) \cite{schadd2009thurn}, \emph{Mr. Jack} (Hurrican, 2006) \cite{mazyad2015jack} as well as card games such as \emph{Lords of War} (Black Box, 2012) \cite{Sephton2014HeuristicMP} and \emph{Magic: the Gathering} (Wizards of the Coast, 1993) \cite{Cowling2012EnsembleDI}. Evolutionary algorithms have been leveraged to play \emph{Splendor} \cite{bravi2019rinascimento} via RHEA and \emph{7 Wonders} (Repos, 2010) \cite{robilliard2016wonders} via genetic programming. AI agents have also been leveraged to playtest and improve the balance of games such as \emph{Ticket to Ride} (Days of Wonder, 2004) \cite{silva2017ticket} and \emph{Dominion} (Rio Grande, 2008) \cite{mahlmann2012dominion}. Finally, of special note is the competitive game \emph{Hanabi} (Abacusspiele, 2010) which is played in teams, and thus AI agents must adapt to their collaborators' mindset \cite{canaan2020hanabi}. 

While preparing this publication, Sauma-Chac{\'o}n and Eger introduced PAIndemic, an A* based agent for playing Pandemic \cite{sauma2020pandemic}. Identifying the need for more collaborative gameplay, the authors developed seven heuristics which were aggregated as a weighted sum to evaluate the state. Interestingly, PAIndemic makes a plan from the current game state without simulating the stochastic game-state changes, and then perform a hundred rollouts on the forward model. PAIndemic is similar to the proposed RHEA agent but does not use macro-actions, and it seems to perform well. It could be worth investigating whether the two methods could be integrated, e.g. using the heuristics of \cite{sauma2020pandemic} for decisions that the RHEA has no control over, such as choosing cards to share or to discard.

\section{Pandemic Game Testbed}\label{sec:game}
Pandemic is a cooperative game that can be played by 2 to 4 players. Players act as a group of scientists who try to save the world from four deadly diseases. Their goal is to cure those diseases, while micro-managing various threats that occur on the map during gameplay. This section summarizes the game rules, initial setup, and way that the game becomes progressively harder (but more predictable).

The game takes place on a simplified version of the world map, which includes 48 cities in the form of a graph, as shown in Fig. \ref{fig:example_moves}. Players can move their pawns from city to city via the available edges between cities. Cities are divided into four colors corresponding to a specific disease type, with 12 cities per color. The version of Pandemic implemented in this paper has the following components: 4 player pawns, 4 role cards (operations expert, researcher, medic and scientist), 48 city cards (of four colors), 6 epidemic cards (although most games reported here use fewer epidemic cards), 48 infection cards (one per city), 96 disease cubes (24 per color), and 6 research station tokens. The tabletop Pandemic game has 3 more player roles and 5 event cards which are mixed with the city cards and allow for special actions played out of turn. In order to simplify the game's state-space, these additional roles (e.g. the dispatcher, who allows a player to move another player's pawn) and event cards (which could allow e.g. `actions' during the infection phase) are not used in this AI testbed.

The game's initial setup is a complex process which also depends on the number of players and the game difficulty. First, nine infection cards are drawn from the shuffled infection deck, and nine cities are infected with a total of 18 disease cubes (3 cubes for the first 3 cities, 2 cubes for the next 3 cities, 1 cube for the last 3 cities). Infection cards drawn in this way are placed in the infection discard pile. In the second step, players are assigned a role and a pawn colored after the role. All players' pawns are placed on Atlanta, together with the first research station. Then players are dealt their initial hands, consisting of city cards. The number of initial city cards per player depends on the number of players (4 cards for two players, 3 cards for three players, 2 cards for four players). Finally, the player cards' deck is prepared: this deck is where players will draw new cards from at the end of their turn. All remaining city cards (not dealt to players) are mixed with a number of epidemic cards in a specific manner (4 epidemic cards for easy, 5 for medium, 6 for hard games). The city cards are equally split into as many sub-stacks as the number of epidemic cards (4, 5 or 6), and one epidemic card is added to each sub-stack. Sub-stacks are individually shuffled and then joined together, one on top of another, without mixing the cards between them. This ensures that epidemics are distributed in the player cards' deck in a uniform manner. After making the player deck, the game begins.

Each player acts in their turn, with the same player order kept throughout the game. On their turn, players can take up to four actions. There are 8 types of actions available to all players, although certain player roles modify the prerequisites and effects of certain actions. Four action types allow players to move around the board, specifically: \textbf{Drive/Ferry} (move along an edge to an adjacent city), \textbf{Direct Flight} (discard a city card to move to the city named on the card), \textbf{Charter Flight} (discard the city card that matches the player's current city to move to any city), \textbf{Shuttle Flight} (move from a city with a research station to any other city with a research station). The other four actions allow players to work towards winning (and against losing) the game. \textbf{Build a Research Station} allows a player to place a research station in the city they are in by discarding a card named after that city. If all 6 research stations have been built, a research station is moved to the current city from anywhere on the board. \textbf{Treat Disease} lets a player remove 1 disease cube from the city they are in, placing it back in the cube supply. If a disease has been cured, this action removes all cubes of that color from the player's current city. \textbf{Share Knowledge} allows a player in the same city with another player to give or take a card from the other player's hand if the card is named after the city. Players with more than 7 cards must immediately discard extra cards. \textbf{Cure Disease} allows a player at any research station to discard 5 city cards of the same color to cure the disease of that color.

Different roles modify these actions. The operations expert role can move from a research station to any city by discarding any city card (once per turn), and can build a research station in their current city without discarding a city card. The researcher role does not need to be in the same city to give a card, as long as the other player is in the city named in the card. The medic role can remove all cubes of the same color with one treat disease action, and if the disease is cured then all cubes of that color in  the city are removed automatically (no need to spend an action). The scientist role must discard only 4 city cards of the same color to cure a disease. Each role has a specialty: the operations expert can build and exploit research stations, the medic can treat disease, the researcher can share knowledge and the scientist can cure diseases. These roles were chosen for this paper due to their clear-cut specialization.

A core design pattern of Pandemic and most collaborative games is the escalating difficulty imposed by the game's rules as the game progresses. In Pandemic, difficulty escalates in several ways which are tightly connected with the core game loop, i.e.: a player takes up to four actions (described above), then draws two cards from the player deck (discarding excess cards above 7 from their hand), then draws a number of cards from the infection deck. The number of infection cards drawn increases as the game progresses (discussed later). Each infection card drawn lists a city: if a city has less than 3 disease cubes of the same color, then a disease cube of a color appropriate to the city is added. If an infection would add a cube in a city that already has 3 cubes of this color, an \emph{outbreak} occurs, the outbreak counter increases, and instead of adding one cube to the city all adjacent cities receive a cube of that color---which may trigger additional outbreaks. 

While outbreaks and infections increase the number of disease cubes on the board, the core way in which difficulty ramps up is through the epidemic card's effects. If an epidemic is drawn from the players' deck, the number of cards drawn in every infection phase from now on may increase (2 infection cards for 0-2 epidemics, 3 cards for 3-4 epidemics, 4 for 5-6 epidemics). Afterwards, the bottom-most card on the infection deck receives 3 disease cubes of that color and is discarded; then, all discarded infection cards are shuffled and placed on top of the deck. Once this is done, the next infection phase will draw infection cards among those previously discarded, ensuring that cubes are placed in cities that were infected before. The epidemic card mechanic increases the difficulty as the same cities keep being infected (which can trigger outbreaks), but it also makes the game ``readable'' to players as they can anticipate which cities will be infected (but not their order) and focus efforts on saving those cities.

The game is won if the players discover cures for all four disease types. The game is lost under any of the following conditions: \textbf{
outbreaks} (if  the outbreaks counter reaches 8), \textbf{disease cubes} (if there are not enough disease cubes to complete an infection step),  \textbf{player cards} (if there are not enough player cards for the player to draw from). The last condition is inevitably met after a specific number of rounds.

It is evident that players risk failure if they neglect either the disease cubes on the board or the number of outbreaks. At the same time players are forced to find cards of the same color (through the share knowledge action, or by drawing them from the deck) in order to cure diseases before the player deck runs out. The balance between pessimistic play (i.e. warding against current threats that may lead to failure) and optimistic play (i.e. taking small steps towards victory) is an important dimension of Pandemic both for humans and for AI agents.

\section{Methodology for AI Game Playing}\label{sec:methodology}

This section discusses the RHEA implementation that handles decision-making of all players in a game of Pandemic, as well as the modifications to the forward model and game-state abstractions necessary to make it work.

\subsection{Game State Representation \& Forward Model}\label{sec:methodology_gamestate}

The game-state is represented as a city graph: each city stores the player pawns, disease cubes, and research stations that exist there. The game-state also stores the current outbreaks and infection rates, as well as each player's cards in their hand. Finally, the game-state stores the discarded and face-down decks for player cards and infection cards, even though the agents' decision-making does not use this additional data; it is only used for the forward model simulations.

Simulating future states in Pandemic is challenging due to the complex way in which the player deck is constructed initially and the way the infection deck changes by recycling discarded cards during play. For the player deck, the forward model keeps track of the size of each sub-stack (see Section \ref{sec:game}) and whether it still contains an epidemic card. To randomize the forward model, all city cards (not epidemics) in the player deck are shuffled together, and sub-stacks of the right size are created from this shuffled global set; after this, sub-stacks still containing epidemics have an epidemic card shuffled into them and the sub-stacks are placed one on top of the other to create the player deck. For the infection deck, after the first epidemic the deck will consist of a number of sub-stacks which contain past discarded cards that have been reshuffled. The game-state keeps track of these sub-stacks and which infection cards are in each; to randomize the forward model, each sub-stack is shuffled individually and then placed one on top of the other in the right order. These steps ensure that the distribution of cards is maintained.

\begin{figure*}
\centering
\includegraphics[trim=0cm 0cm 0cm 4cm,clip=true,width=0.95\textwidth]{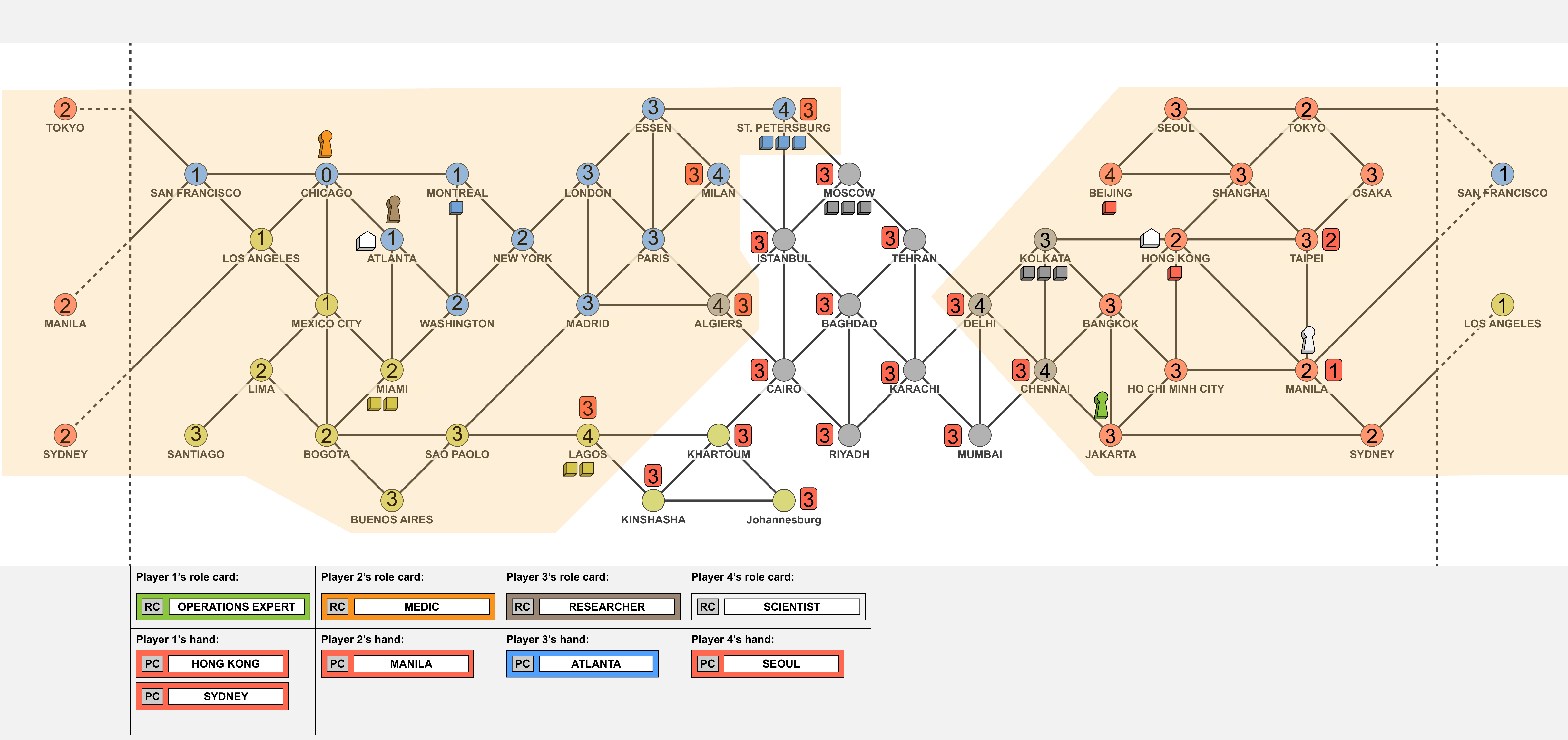}
\caption{A view of the Pandemic board, highlighting the areas accessible to Player~2 (P2, in orange) with the different move options (measuring the shortest route). P2 can move without spending a card to the orange highlighted areas, spending a number of actions shown inside each city. P2 can take a shuttle flight from Atlanta to Hong Kong (both have research stations). P2 can spend the Manila card to travel faster to Manila and Taipei via a direct flight from Chicago. P2 can also travel to Manila via drive/ferry and spend the Manila card there (using the charter flight) to travel anywhere in the world. All cities accessible by spending the red Manila card are shown next to the city with the least actions spent inside a red rectangle.}
\label{fig:example_moves}
\end{figure*}

\subsection{Action Representation}\label{sec:methodology_actions}

Since players have four actions per turn and can move in different ways around the board, the combination of possible actions is prohibitive for an efficient AI agent. Instead, the action space for Pandemic agents is represented via macro-actions, i.e. a sequence of one or more actions which includes movement actions for getting to the required location and the action taken at that location. The system produces all the available macro-actions in a compositional manner. It first composes all the possible movement sequences, including simple movement as well as movement using special abilities and cards. Afterwards, it selects the movement sequences that can lead the player to a specific non-movement action on the map. The final product of this process is an almost complete set of all the possible courses of action that the player could take and that would make sense to consider. The following macro-actions are considered:

\begin{itemize}
    \item \textbf{Treat disease macro-actions:} the system finds all cities with one or more disease cubes and finds the optimal path of $[0,N-1]$ move actions to reach it, followed by a treat disease action at that city.
    \item \textbf{Cure disease macro-actions:} if the current player has enough cards to cure a disease (5 cards, or 4 for Scientist) then the macro-action finds the optimal path of $[0,N-1]$ move actions to the closest research station, followed by a cure disease action at that city.
    \item \textbf{Build research station macro-actions:} the system first finds all cities in which the player can build a research station: for the Operations Expert that would be any city that does not already have a research station, and for other players it would be a city for which they have a card in their hand. The macro-action then finds the optimal path of $[0,N-1]$ move actions to all candidate cities (as long as those moves do not cost the city card in question), followed by a build action at that city.
    \item \textbf{Share knowledge macro-actions:} this is a complex system which finds all cities where the player can give or take cards, or to wait there for another player (so that they can exchange cards later). The system first scans the players' hands, in order to create a list of all the possible card exchanges between players. For each exchange, it finds the city where the exchange must take place. For every exchange location, it then finds all the ways that the current player can get there. Finally, it combines move actions with the following share knowledge action, or a `wait' action if the other player is not yet in that location.
    \item \textbf{``Walk away'' macro-actions:} this special set of macro-actions only consists of $N$ move actions with no end-goal. Any city that is exactly $N$ move actions from the current player's city is chosen, and the player moves to that city. 
    These macros consist of move sequences of any type, and may even spend ``valuable'' cards. However, these macros are rarely used as it is quite improbable that none of the other macro-actions are available.
\end{itemize}

Each of these macro-actions can include any number of movement actions calculated based on the shortest action sequence (see Fig.~\ref{fig:example_moves}). Eligible movement actions include any drive/ferry action and shuttle flight action (as they do not require spending cards) and any direct flight and charter flight for which the card spent does not reduce the overall chances of curing a disease. The same metric is used to choose cards to discard in case the player has more than 7 cards, and for selecting cards to give or take through the share knowledge macro-actions. For any disease $t$, the ability to cure the disease is measured via $A(t)$ in Eq.~\eqref{eq:ability_cure_all} which depends on the best hand across all players (in terms of cards of this type).
\begin{align}
A(t) &=
\begin{cases}
1 &\text{if $t$ cured}\\
max_{p=1{\ldots}P}A_c(p,t) &\text{otherwise}
\end{cases}\label{eq:ability_cure_all}\\
A_c(p,t) &=
\begin{cases}
1 &\text{if ${h(p,t)\geq}C_d(p)$}\\
\frac{h(p,t)}{C_d(p)} &\text{otherwise}
\end{cases}\label{eq:ability_cure_individual}
\end{align}
\noindent where $P$ is the number of players, $h(p,t)$ is the number of cards of type $t$ in the hand of player $p$ and $C_d(p)$ is the number of cards needed for player $p$ to cure a disease ($C_d=4$ for the Scientist, and $C_d=5$ for every other role). $A_c(p,t)$ of Eq.~\eqref{eq:ability_cure_individual} captures each player's ability to cure disease $t$.

Note that in the above descriptions, $N$ refers to the number of actions that the player wishes to ``invest'' for such a macro-action. In experiments reported in this paper, we consider only the macro-action that can be completed within a player's current turn, i.e. with $N$ equal or lower the player's remaining actions (max. 4, if the player has not taken any actions yet).

\subsection{RHEA for Pandemic}\label{sec:methodology_rhea}

The RHEA implemented in this paper applies a number of mutations on an initial seed produced by a hand-crafted `hierarchical policy' agent (HPA), discussed in Section \ref{sec:methodology_rhea_init}. This HPA has a small degree of randomness but largely follows a carefully crafted order of macro-actions, only choosing low-priority ones if no high-priority macro-actions are possible.

The RHEA for Pandemic follows a 1+1 evolutionary strategy with the following steps:
\begin{enumerate}
    \item Initialization: an individual is generated via HPA
    \item Evaluation: the individual's macro-actions are simulated in one or more instances of the forward model and the final state(s) evaluated via a selected heuristic.
    \item Mutation: the individual is copied and mutated, to form the offspring which is also evaluated.
    \item Replacement: If the offspring is better than its parent, it replaces the parent. Otherwise it is discarded.
    \item After a number of iterations, the process stops and the algorithm returns the first action of the current individual. Otherwise, the algorithm repeats steps 3-5.    
\end{enumerate}

\subsubsection{Genetic encoding}\label{sec:methodology_rhea_genetic_encoding}
Each individual consists of genes that are neither single actions nor macro actions, but a higher-level construct that represents a set of macro actions that may occur within the span of a player's turn. Every individual consists of an exact number of such genes, equal to the agent's look-ahead ($H$), measured in player turns. Most experiments in this paper use 5 genes per individual ($H=5$).

\subsubsection{Initialization}\label{sec:methodology_rhea_init}
The initial seed of the RHEA is generated by the hierarchical policy agent (HPA). The HPA enumerates all possible macro-actions of a specific type (based on a predefined order): if there are any macro-actions of this type then a random one of them is chosen and executed, otherwise the next type of macro-actions is enumerated etc. The following order for HPA was chosen following intuition and experimentation:
\begin{enumerate}
\item Cure disease macro-actions
\item Treat disease macro-actions only for cities with 3 disease cubes of the same type
\item Share knowledge macro-actions (take or give) with immediate effect, otherwise wait in position to share knowledge on another player's turn (take or give)
\item Build research station macro-actions (if there are less than 6 research stations)
\item Treat disease macro-actions only for cities with 2 disease cubes of the same type
\item Treat disease macro-actions only for cities with 1 disease cube of the same type
\item Walk away (i.e. move randomly using all remaining actions left for this player's turn)
\end{enumerate}

\subsubsection{Genetic operators}\label{sec:methodology_rhea_operators}

The mutation process is implemented as a partial destruction and stochastic repair, described below.
The algorithm iterates over the genes of the current individual. A fixed ``mutation rate'' determines the probability that a gene will be mutated. If a gene is selected for mutation, a macro action from within that gene (i.e. player's turn) is randomly selected. This macro action, as well as its consequent ones (within that gene) are deleted, while the previous ones remain intact. 
Once the selected gene has been (partially) destroyed, the repair process takes place: a randomized game state is rolled forward, by using the macro-actions that are inscribed in the genome, up to the point of destruction. During this forward-rolling of the state, there may be macro actions that are not applicable due to the stochastic nature of the forward model. In that case, all incompatible actions are ignored: e.g. when performing a macro-action to treat disease in London, but London in this rollout has no disease cubes, the player performs the move actions to get to London but ``wastes'' the treat disease action at the end. 
As soon as the state is rolled to the point of repair, a Random-order Policy Agent (RPA) selects the next macro action, while the HPA instantiates any remaining actions until the gene is fully repaired. When this process is over, the algorithm continues to iterate over the individual's genes, possibly applying more mutations. If during the iteration no gene was selected, then one gene is selected at random and mutated as above.

The RPA randomizes the order of the following types of macro-actions. Once an order is set, RPA finds all possible macro-actions of that type and selects a random one; if none exists, it checks the next type of macro-actions etc.
\begin{enumerate}
\item Cure disease macro-actions
\item Treat disease macro-actions only for cities with 3 disease cubes of the same type; if none exist, treat disease macro-actions for cities with 2 disease cubes of the same type; if none exist, treat disease macro-actions for cities with 1 disease cubes of the same type.
\item Share knowledge macro-actions (take or give) with immediate effect, otherwise wait in position to share knowledge on another player's turn (take or give)
\item Build research station macro-actions (if there are less than 6 research stations)
\end{enumerate}

\subsubsection{Fitness definitions}\label{sec:methodology_rhea_fitness}
Based on the end-game conditions of {Pandemic} (see Section \ref{sec:game}), there are several ways to evaluate any given state: \emph{optimistically} in terms of the cards needed to discover every cure, or \emph{pessimistically} in terms of the disease cubes or outbreaks left before the game is lost.
The following state evaluation (fitness) functions are tested in this paper:
\begin{align}
f_{o,d}&=\frac{1}{4}N_d \label{eq:fod}\\
f_{o,A}&=\frac{1}{1.3}\left( \frac{1}{4}\sum_{t=1}^4A(t)+0.3{\cdot}N_d\right) \label{eq:foa}\\
f_{c,a}&=\frac{1}{4}\sum_{t=1}^4\frac{N_{c}(t)}{24} \label{eq:fca}\\
f_{c,m}&=min_{t=1{\ldots}4}\frac{N_{c}(t)}{24} \label{eq:fcm}\\
f_{c,p}&=\prod_{t=1}^4\frac{N_{c}(t)}{24} \label{eq:fcp}\\
f_{b}&=1-\frac{N_b}{8} \label{eq:fb}
\end{align}
\noindent where $N_d$ is the number of cured diseases, $N_{c}(t)$ is the number of cubes for disease $t$ remaining off the board, and $N_b$ is the number of outbreaks that have occurred so far.

The fitness functions account for cured diseases ($f_{o,d}$) or the general ability to cure diseases ($f_{o,a}$), different ways to calculate disease cubes remaining off the board (average, minimum, or product) and finally the number of outbreaks (as the game ends at 8 outbreaks). All fitness scores are normalized to $[0,1]$ and a high fitness indicates a better game state. Of note is the addition of $0.3{\cdot}N_d$ in Eq.~\eqref{eq:foa} which gives additional pressure if the disease is already cured compared to instances where the disease \emph{can} be cured.

\section{Experiment}\label{sec:experiment}

\begin{figure}
    \centering
    \includegraphics[width=0.42\textwidth]{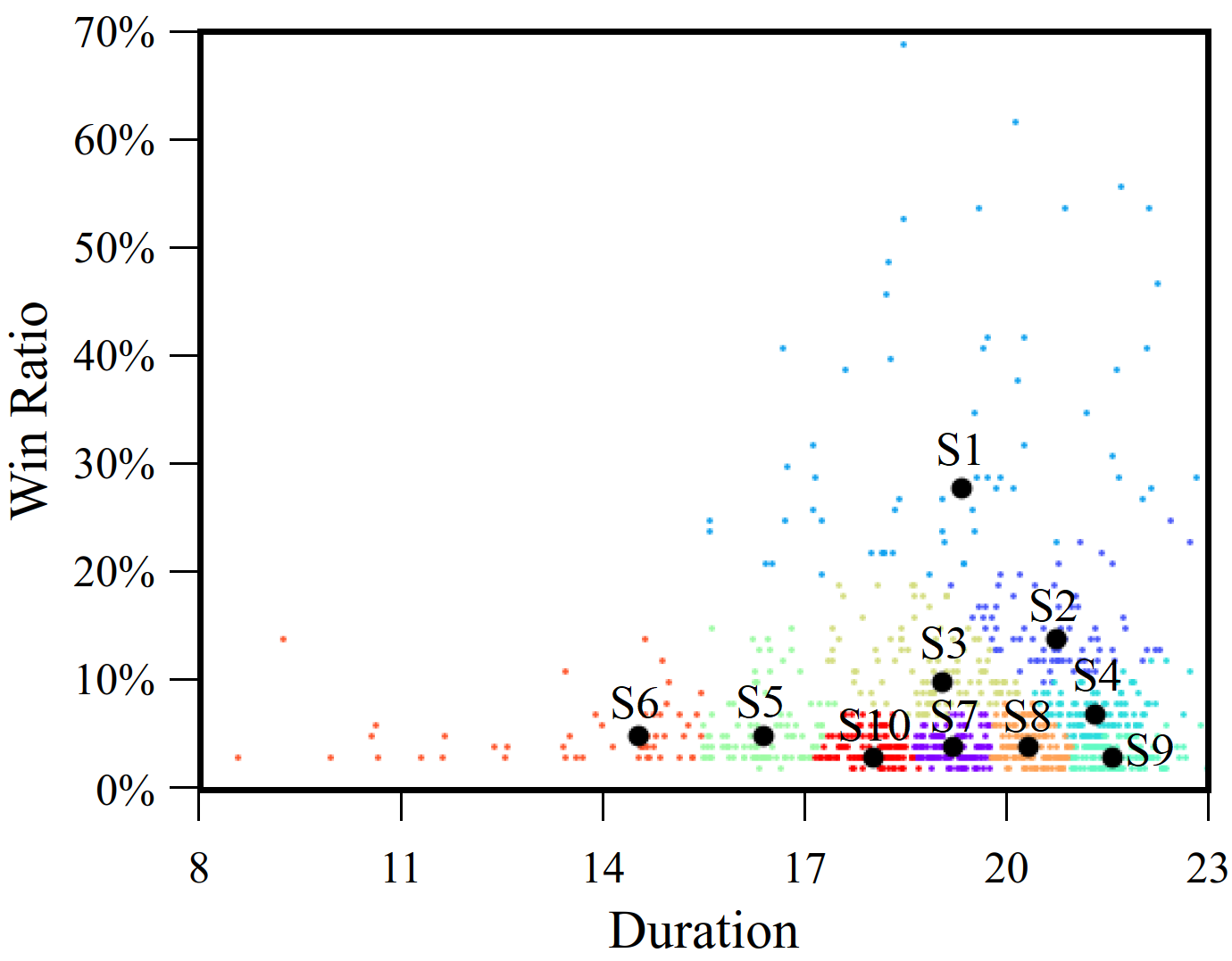}
    \caption{Ten testbed setups (black dots) via $k$-medoids clustering on the 1000 `easiest' setups based on HPA performance.}
    \label{fig:clustering}
\end{figure}

To assess the performance of the RHEA agent which controls all player's actions, a deterministic environment is ideal. However, Pandemic includes many stochastic elements during game setup and when infection cards are reshuffled during the game due to epidemic cards. Experiments in this paper control for the former (initial game setup) by using 10 fixed game starts (including players' roles and order of play, player decks and infection decks) although resolving epidemics during the game will still introduce some stochasticity. The HPA, which is used as a baseline throughout this paper, is used to find representative game setups that use 4 epidemic cards (easy difficulty) and the same four player roles in the same turn order: (1) Operations expert, (2) Medic, (3) Researcher, (4) Scientist. A set of $10^4$ random initial setups are tested by the HPA, and the  $10^3$ easiest ones are chosen (i.e. where the HPA has a higher win ratio in 100 trials of each game setup). To select a smaller but representative set from these $10^3$ setups, ten initial setups were selected as the medoids from clustering along the axes of win ratio (naturally between 0 and 1) and duration (normalized based on the maximum game length, i.e. 23 turns). The distribution of the top $10^3$ setups and the ten medoids are shown in Fig. \ref{fig:clustering}. The average win ratio for the hierarchical policy agent for these setups is 8.3\% (ranging from 28\% to 3\%), and an average game duration of 19 turns (ranging from 13.9 to 20.6 turns). 

All experiments except Section \ref{sec:experiments_robustness} test these ten setups with the same order of unseen cards, easy difficulty, and player order as discussed above. Each setup is played until won or lost for 100 runs. Performance metrics of note is the win ratio in 100 runs, as well as improvement of RHEA in terms of win ratio over the baseline HPA. 

\subsection{Impact of State Evaluation}\label{sec:experiments_evaluation}

A number of fitnesses are proposed in Section \ref{sec:methodology_rhea_fitness} for evaluating the state of the game: Eq.~\eqref{eq:fod}-\eqref{eq:foa} are optimistic (taking into account how ``close'' the game is to being won) and Eq.~\eqref{eq:fca}-\eqref{eq:fb} are pessimistic (taking into account how ``far'' the game is to being lost). These fitnesses do not inherently consider whether the game is already won or lost. Variations of each fitness are also tested: Eq.~\eqref{eq:winlose} assigns maximum fitness when the game is won and minimum fitness when the game is lost, while Eq.~\eqref{eq:penalty} rewards winning in the same way but penalizes losing proportionately to the fitness score. The $p(f)$ formula hypothesizes that while losing should always be penalized compared to staying in the game (via the $C_p$ constant), the game-state when losing can indicate how well the agent could defend against a loss. In this paper $C_p=0.1$. 
\begin{align}
w(f)&=
\begin{cases}
1 &\text{if game won}\\
f &\text{if game ongoing}\\
0 &\text{if game lost}
\end{cases}\label{eq:winlose}\\
p(f)&=
\begin{cases}
1 &\text{if game won}\\
f &\text{if game ongoing}\\
C_p{\cdot}f &\text{if game lost}
\end{cases}\label{eq:penalty}
\end{align}
Reported experiments evolve the agent for 100 generations and fitness is averaged from 5 rollouts of the macro-action sequence using a different stochastic forward model each time.

\subsubsection{Single evaluation}\label{sec:experiments_evaluation_single}

\begin{figure}
\centering
\subfloat[State evaluation as single fitness]{
\includegraphics[trim=0mm 2mm 0mm 2mm, clip,height=0.2\textwidth]{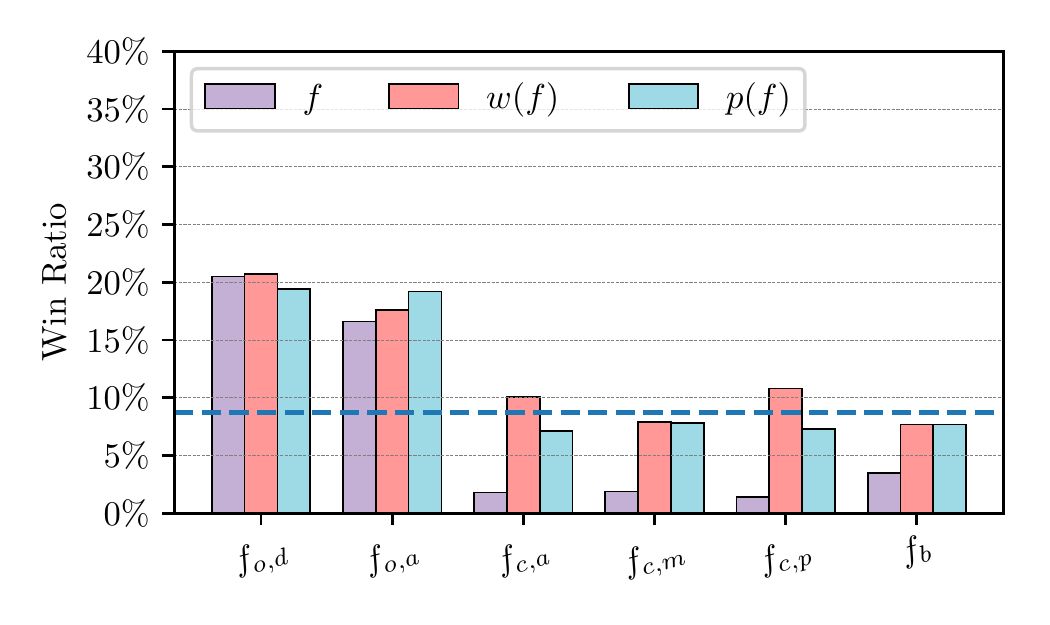}
\label{fig:experiments_evaluation_single} 
}\\
\subfloat[State evaluation as average fitness of the two fitness scores]{
\includegraphics[trim=0mm 1mm 0mm 2mm, clip,height=0.2\textwidth]{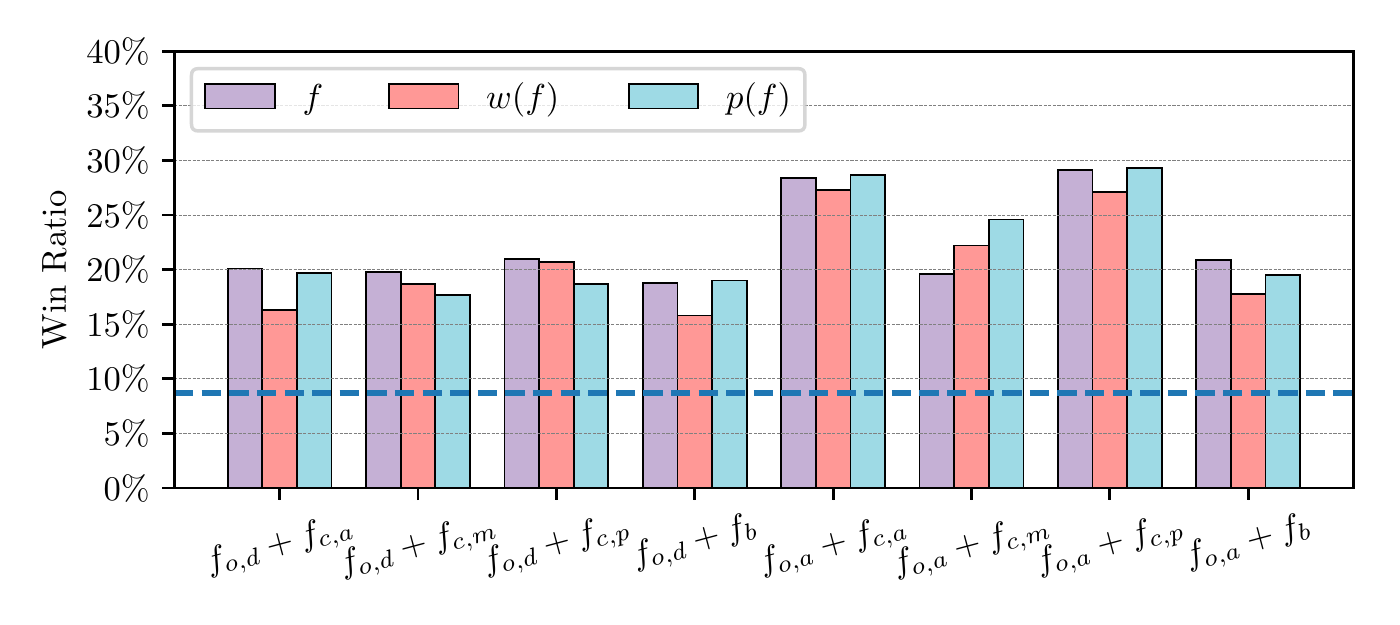}
\label{fig:experiments_evaluation_aggregated}
}
\caption{Average win ratio for the 10 test setups, using different state evaluations (with or without win/loss conditions). The dotted line is the win ratio of the hand-crafted HPA.}\label{fig:experiments_evaluation} 
\end{figure}

This experiment tests each fitness of Section \ref{sec:methodology_rhea_fitness} in its three variants: the average win ratio in the ten testbed setups (from 100 trials in each setup) are reported in Fig.~\ref{fig:experiments_evaluation_single}. An important observation is that pessimistic evaluations on their own perform much worse than the HPA, or comparably when winning and losing conditions are accounted for. Generally, the $w(f)$ variant performs better than the unprocessed fitness, while $p(f)$ outperforms $w(f)$ only for $f_{o,a}$. The optimistic fitnesses manage to steer the agent towards winning the game more often: the best improvement over the baseline (averaged across the ten setups) is 120\% with $w(f_{o,d})$. In terms of other differences between the agents, optimistic agents generally tend to play shorter games and lose much faster than pessimistic agents. Indicatively, $f_{o,d}$ is the fastest to lose, with lost games' average duration at 14.3 player turns; in contrast, $f_{c,p}$ is the slowest to lose (21.3 turns). Unsurprisingly, pessimistic agents who prioritize keeping disease cubes off the board rarely lose due to outbreaks or insufficient cubes: indicatively, for $f_{d,p}$ only 22\% of lost games are due to epidemics and 6.3\% due to disease cubes, compared to 56\% and 24\% respectively for $f_{o,a}$. Finally, optimistic agents tend to use the share knowledge action more often than the HPA while the opposite is true for pessimistic agents. Pessimistic agents tend to use the treat disease action more often than the HPA, while the opposite is true for optimistic agents. These differences in actions favored are less pronounced when the fitness is conditionally applied as $w(f)$ or $p(f)$.

\subsubsection{Combined evaluation}\label{sec:experiments_evaluation_aggregated}

While fitness functions measuring how close players are to winning seem to perform well, optimistic RHEA agents underestimate losing conditions and tend to lose quickly. The hypothesis is that combining optimistic and pessimistic fitnesses could allow agents to account for both opportunities and dangers in their final state. For the sake of this experiment, two fitness scores are averaged (one optimistic, one pessimistic) and applied either on their own or conditionally via Eq.~\eqref{eq:winlose} and Eq.~\eqref{eq:penalty}. 

The average win ratio for different combinations of state evaluations, from 100 trials per setup, are reported in Fig.~\ref{fig:experiments_evaluation_aggregated}. 
Unlike in Fig.~\ref{fig:experiments_evaluation_single}, the naive aggregated state evaluation often performs better than the conditional variants, especially $w(f)$. While $f_{o,d}$ performed better on average than $f_{o,a}$ when applied alone, in this case fitnesses that combine $f_{o,a}$ perform much better. While differences are quite small among the most well-performing agents, the best agent is $p(\tfrac{f_{o,a}+f_{c,m}}{2})$ with an average win ratio of 29.3\%, (302\% improvement over HPA). 

\subsection{Impact of Mutation Probabilities}\label{sec:experiments_mutation}

Based on the experiments of Section \ref{sec:experiments_evaluation}, the best agent used $p(\tfrac{f_{o,a}+f_{c,m}}{2})$ as its state evaluation. It should be noted however that all experiments in Section \ref{sec:experiments_evaluation} use a mutation probability of 50\%; this section explores instead how mutation chances may impact the performance of the best RH agent.
Three constant mutation probabilities are tested: 100\%, 50\% and 0\%. With 100\% mutation one macro-action is replaced with one produced by the RPA in every player's turn, while with 0\% chance only one player's action will be mutated (triggering the failsafe as discussed in Section \ref{sec:methodology_rhea_operators}). In addition, we test three variants where the mutation probability drops during the course of evolution: $100\% \rightarrow 50\%$, $75\% \rightarrow 25\%$ and $50\% \rightarrow 0\%$. The mutation rate starts at the left-most value and decreases by a small increment per generation, until at the end of evolution it reaches its right-most value.

\begin{table}
\centering
\begin{tabular}{|c|c|c|c|}
\hline
Constant & $MR_{0\%}$  &  $MR_{50\%}$ & $MR_{100\%}$  \\ \hline
Win Rate & 19.5\% & 31.4\% & 34.9\% \\\hline\hline
Dropping & $MR_{50\% \rightarrow 0\%}$ & $MR_{75\% \rightarrow 25\%}$ & $MR_{100\% \rightarrow 50\%}$  \\ \hline
Win Rate & 25.6\% & 31.4\% & 35.1\% \\ \hline
\end{tabular}
\caption{Win rates for the agent with the best game-state evaluation, when the mutation rate is constant or varies, averaged across the 10 testbed games.}
\label{tab:mr_type_of_result}
\end{table}

As shown in Table \ref{tab:mr_type_of_result}, the highest mutation rates ($MR_{100\%}$ and $MR_{100\% \rightarrow 50\%}$) result in a higher win ratio at $\sim$35\%. The worst performing agent was with 0\% mutation rate at $19\%$ win ratio. This observation may seem counter-intuitive at first glance, as slow mutations were expected to more smoothly guide evolution towards a solution, especially in this highly stochastic simulation. 

A possible explanation for the improved behavior at high mutation rates is that certain parts of the gameplay require more than one round in order to be properly executed. Share knowledge actions are an obvious example. If one player's turn is mutated to go to a city and wait to share knowledge but the next player's action is not mutated to go to the same city and complete the share knowledge action, the first player's actions are wasted. Similarly, if a mutation leads to a new research station being built, some cities become easier to reach for all players. However, if other players' actions are not mutated then the system will not re-calculate the shortest path with the new research station node. It seems that high mutation rates are better at supporting both types of behaviors. Both $MR_{100\%}$ and $MR_{100\% \rightarrow 50\%}$ have a higher rate of {share knowledge} and {build research station} actions per turn. Specifically, $MR_{100\% \rightarrow 50\%}$ takes a share knowledge 32\% more often than the worst performing $MR_{0\%}$ and 56\% more often than the HPA, while the build research station action is performed 17\% more often than both $MR_{0\%}$ and HPA.

\subsection{Impact of Computational Resources} \label{sec:experiments_resources}

The {number of generations} and the {number of repeated rollouts for evaluation} are two key parameters that can directly affect both its performance (win ratio) and its required computational resources. The former dictates how many ``chances'' the agent is given to find a better solution. With more generations, the probability of finding better solutions should increase.
Given the stochastic nature of the forward model, on the other hand, more repetitions increase the accuracy of its evaluation. While other experiments kept the number of generations and evaluations constant at 100 and 5 respectively, this section explores their impact on performance and computation time. This section uses the best RHEA agent found so far, with $p(\tfrac{f_{o,a}+f_{c,m}}{2})$ and $MR_{100\% \rightarrow 50\%}$.

Table \ref{tab:performance_heatmap} shows how the best RHEA agent's win ratio fluctuates with the {number of generations} and {number of evaluation repetitions}. Unsurprisingly, increasing either---or both---of the parameters consistently leads to an improved performance. The best win ratio ($52.6\%$) is achieved with 200 generations and 80 evaluation repetitions. To allow us to evaluate the computational cost of raising the agent's performance through the examined parameters, Table \ref{tab:performance_heatmap} includes the computation time per decision, calculated on a mid-tier desktop computer running each evolution variant on a separate core (Intel i7-8700 at 3.2GHz, 6-core CPU, 8GB RAM). Notably, the best parameter pair of (200, 80) takes an average of 2.96 seconds per decision, while the parameter-pair of (100, 5) yields a performance of $32.4\%$ at only $0.51$ seconds per decision. In this instance, for a relative performance gain of 62\%, the operational time increases almost six-fold.

\begin{table}[t]
\centering
\begin{tabular}{|c|c|c|c|c|c|c|}
\cline{2-7}
\multicolumn{1}{c}{} & \multicolumn{6}{|c|}{Evaluation Repetitions} \\ \hline
{Gen.} & 1	&	5	&	10	&	20	&	40  &   80\\ \hline
25 &	11.7\%	&	18.2\%	&	19.5\%	&	22.6\%	&	24.3\%  &   30.1\% \\ 
 &(0.09s) &(0.12s) &(0.13s) &(0.16s) &(0.32s) &(0.45s) \\ \hline
50 &	17.1\%	&	27.1\%	&	30.2\%	&	28.3\%	&	35.1\%  &   40.0\% \\ 
 &(0.17s) &(0.31s) &(0.30s) &(0.41s) &(0.55s) &(0.78s)    \\ \hline
100 &	23.8\%	&	32.4\%	&	38.9\%	&	39.5\%	&	43.7\%  &   44.7\% \\ 
 &(0.34s) &(0.51s) &(0.59s) &(0.68s) &(0.95s) &(1.47s)    \\ \hline
200 &	28.2\%	&	39.5\%	&	44.1\%	&	48.7\%	&	48.1\%  &   52.6\% \\ 
 &(0.73s) &(0.92s) &(1.07s) &(1.36s) &(1.84s) &(2.96s) \\ \hline
\end{tabular}
\caption{Win ratio for different numbers of generations and evaluation repetitions, with the time per decision in seconds shown in parentheses.}
\label{tab:performance_heatmap}
\end{table}

The analysis in this section reaches an expected conclusion, i.e. that the performance of the RHEA can improve beyond what is reported in this paper if we provide it with more computational resources. Experiments have not shown that the win ratio plateaued, although it is expected that increasing the win ratio will come at ever-greater computational effort. The diminishing returns in terms of performance make the choice of an appropriate parameter pairing case-specific. Since the agent needs to be tested in large-scale experiments, we identify that 100 generations and 10 repetitions strike a good balance, reaching a 20\% relative performance improvement than the (100, 5) parameter pair with a 16\% increase in decision time.

\subsection{Impact of Horizon}\label{sec:experiments_horizon}

In its original implementation \cite{perez2013rhea}, the RHEA operates on sequences of actions or macro-actions of a specific length. In our implementation, as explained in \ref{sec:methodology_rhea_genetic_encoding}, the number of macro actions is relatively dynamic, as every gene contains the macro actions of a complete turn. Nevertheless, in experiments thus far the number of genes (horizon) remained fixed and expresses how far ahead (in player turns) the agent can plan. Experiments thus far used a horizon of 5 player turns ($H=5$) based on preliminary testing and the intuition that for a four-player game the current player should consider at least every other player's future turn as well as their own future turn when making a decision. The assumption was that decreasing the horizon to fewer player turns would prevent the agent from devising complex strategies, while increasing it to longer horizons would render its predictions inaccurate (due to stochasticity of the forward models), or require much more computational effort. The current experiment tests this assumption by measuring the agent's win ratio on different horizon lengths. Furthermore, it examines how the number of evaluation repetitions impacts performance when the horizon varies. This experiment uses the best fitness from Section \ref{sec:experiments_evaluation} and the best mutation probability ($MR_{100\% \rightarrow 50\%}$).

\begin{figure}[t]
\centering
\subfloat[10 evaluation repetitions]{
\includegraphics[trim=0mm 3mm 0mm 2mm, clip,width=0.95\columnwidth]{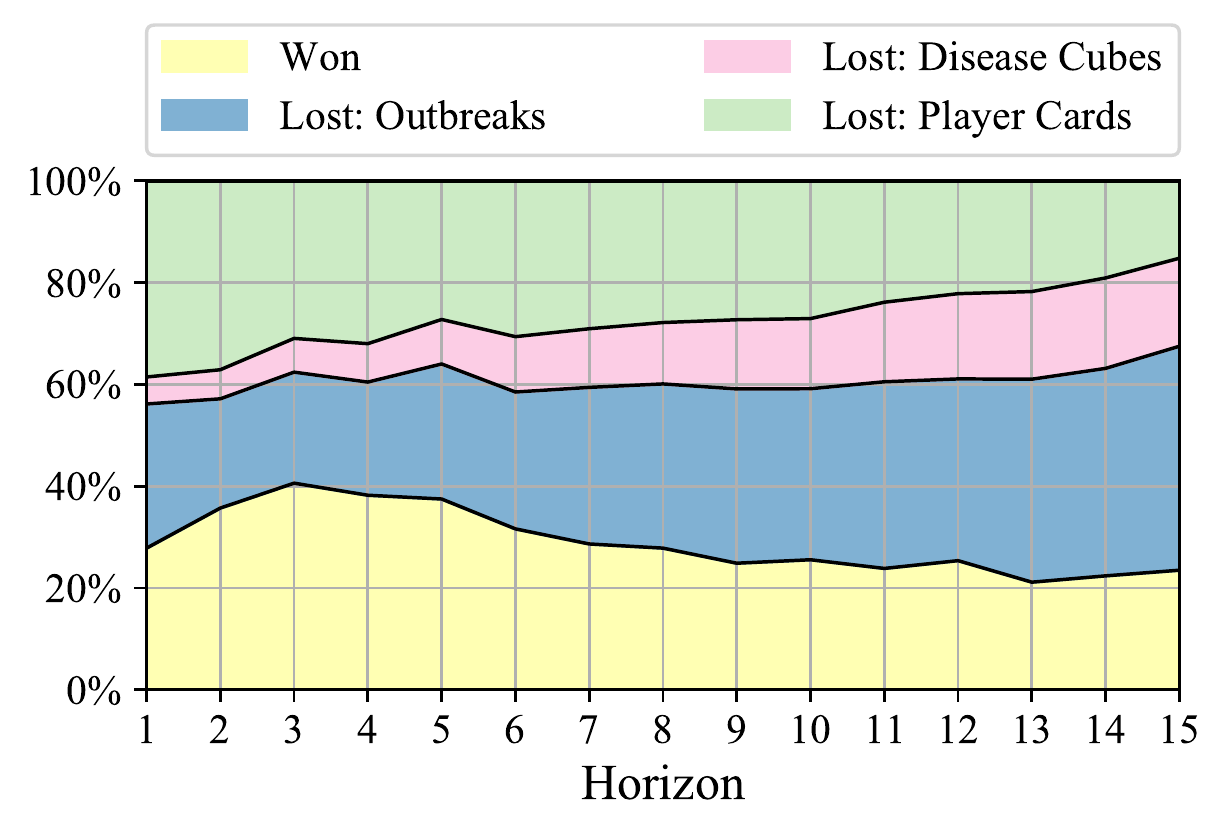}
\label{fig:individual_length_1}
}\\
\subfloat[80 evaluation repetitions]{
\includegraphics[trim=0mm 3mm 0mm 2mm, clip,width=0.95\columnwidth]{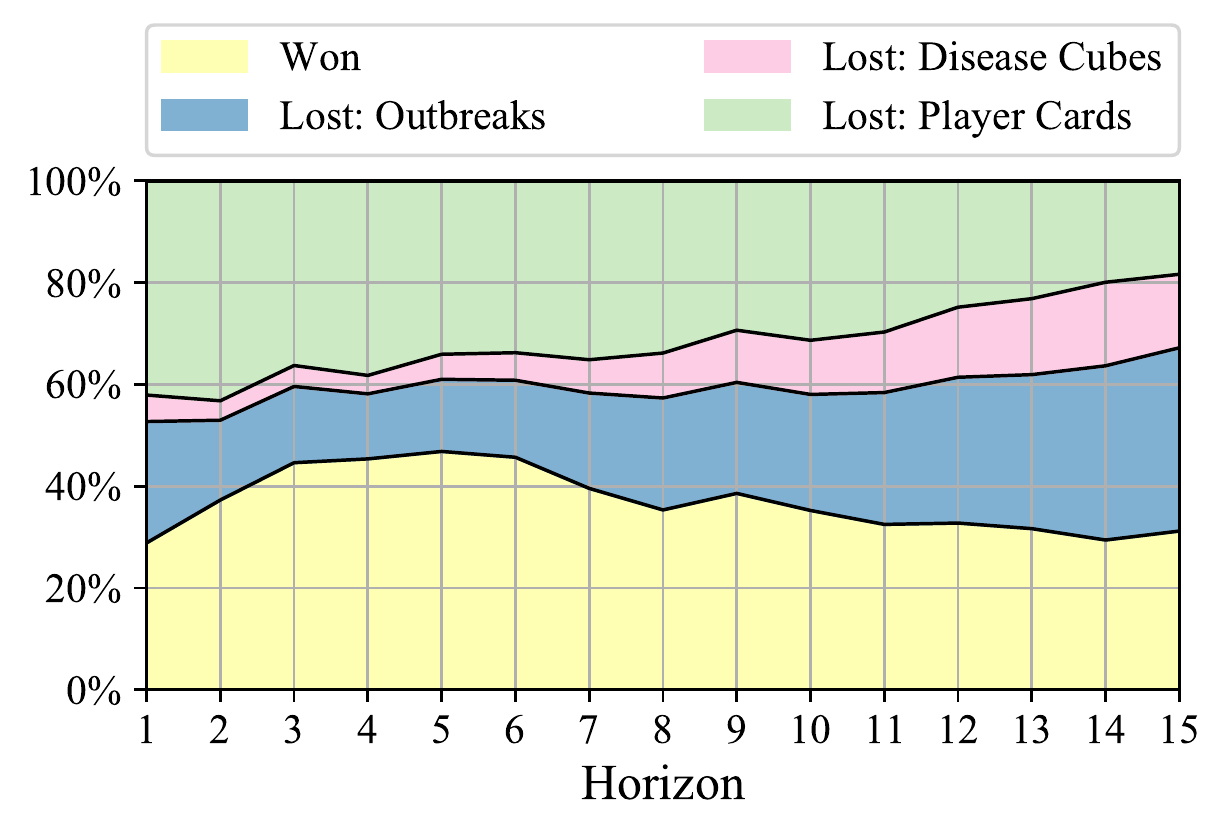}
\label{fig:individual_length_2}
}
\caption{Ratios of each end-game outcome for different horizons (i.e. number of player turns considered), after 100 generations and with different numbers of evaluation repetitions.}
\end{figure}

Figure \ref{fig:individual_length_1} shows the distribution of game ending conditions for various horizon lengths, at 10 evaluation repetitions and 100 generations. As results suggest, the optimal horizon length seems to be 3 player turns (40.6\% win rate), with 4 and 5 turns performing marginally worse ($\sim$38\%). Increasing the horizon beyond 5 turns leads to a drop in performance and the losses due to outbreaks or disease cubes increase dramatically. This suggests that for long horizons, the agent loses its ability to play defensively and manage the upcoming risks.
Since assessing risk can be done better if the agent has more repetitions of the rollouts for state evaluation, Figure \ref{fig:individual_length_2} shows how the RHEA performs in the case of 80 evaluation repetitions and 100 generations. As the results suggest, however, the best performance remains at 5 rounds (47\% win rate), with $H=6$ and $H=4$ coming very close. Notably, the performance drop at longer horizons is less steep compared to Fig.~\ref{fig:individual_length_1}: for $H=15$ performance drops by 42\% (relative to best performance across $H$) at 10 evaluation repetitions versus a drop of 33\% at 80 repetitions. Looking at the distribution of lost games, it seems that at $H\in[3,5]$ the version with 80 repetition primarily loses due to the player deck running out, and manages to defend better against outbreaks (27\% of lost games were due to outbreaks for 80 repetitions at $H=5$, versus 37\% for 10 repetitions at $H=3$).

The main conclusion from these observations is that our initial intuition was not far from the actual optimal value, given the available computational resources. Furthermore, as RHEA struggles to plan many turns ahead, it is evident that the stochastic environment hinders long-term strategies. As an exemplar collaborative game, Pandemic shows the types of challenges that games of this type can pose to AI more broadly.

\subsection{Robustness of performance in random game setups}\label{sec:experiments_robustness}

\begin{figure}
    \centering
    \includegraphics[trim=0mm 3mm 0mm 2mm, clip,width=0.95\columnwidth]{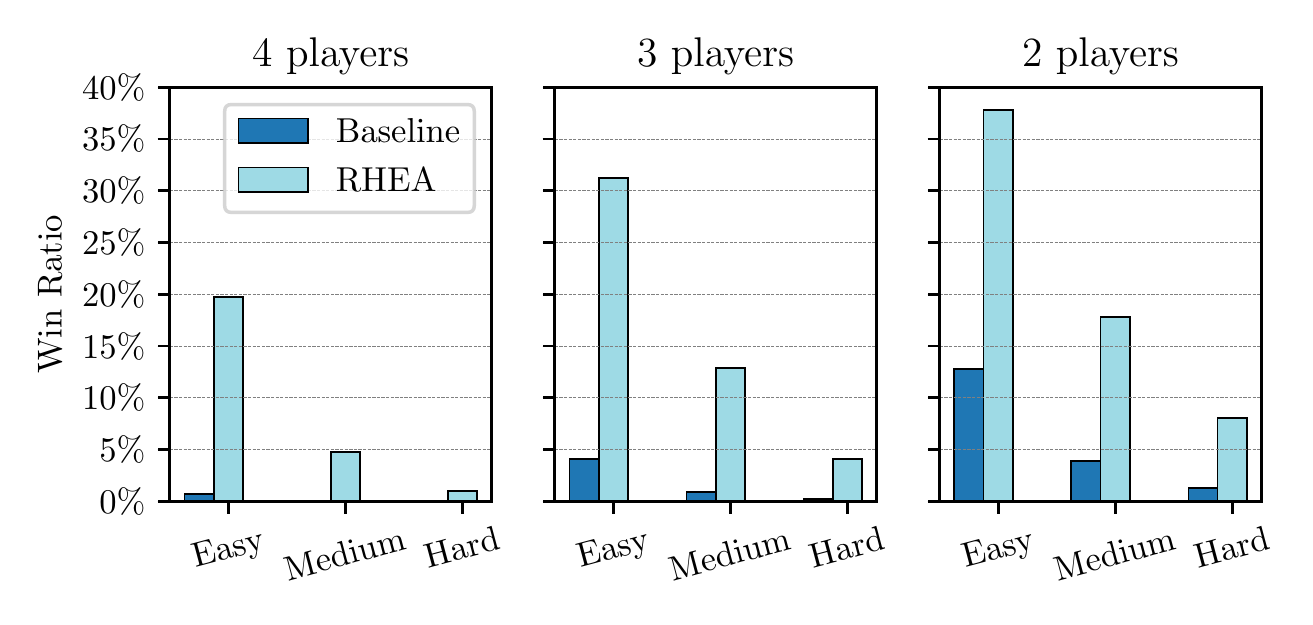}
    \caption{Win ratio of the baseline HPA and the best RHEA agent in $10^4$ games with random roles and random initial game setups, for different number of players and difficulty levels.}
    \label{fig:robustness}
\end{figure}

Based on the previous experiments, the best performing agent uses $p(\tfrac{f_{o,a}+f_{c,m}}{2})$ as state evaluation, a variable mutation rate $MR_{100{\rightarrow}50}$, and a horizon of 3 turns. Based on realistic computational demands, we have concluded on an evolutionary setup of 100 generations and 10 simulations for assessing each individual. This setup results in a 40.6\% win rate (see Fig.~\ref{fig:individual_length_1}) in the carefully selected set of ten initial game states on which parameter tuning took place. This section instead tests how the RHEA fares in a broad set of games, difficulty settings, numbers of players, player roles and turn order. 

Nine experiments are performed, for three different difficulty levels (four, five or six epidemics included in the player deck) and with two, three or four players. The number of players is expected to affect the game's difficulty, because it is easier to coordinate with only one other player and because the game state does not change as drastically (e.g. from infections) between consecutive turns of the same player. Note that for fewer players, the initial players' hand is larger. Moreover, in every game the player roles allocated is randomized: this means that for two- or three-player games some roles will not be present and in all games the turn order is different---unlike the testbed setups where the roles' order was always the same. For each experiment, $10^4$ random games are created and played once by the best RHEA agent (described above) and the baseline HPA agent. The average win ratio for each experiment is presented in Figure \ref{fig:robustness}.

As expected, the win rate of both RHEA and HPA is higher with fewer players across game difficulty levels. The number of epidemics (i.e. game difficulty) has a negative correlation on the win rate of both algorithms, as does the number of players. However, the number of epidemics has a more severe impact on the RHEA (Pearson's $\rho=-0.86$; $p<0.05$) than on the HPA ($\rho=-0.57$; $p>0.05$), while the number of players has a more severe impact on the HPA ($\rho=-0.605$; $p<0.05$) than on the RHEA ($\rho=-0.43$; $p>0.05$). With two players, the RHEA reaches win rates 3 to 6 times the respective win rates of the HPA. For three players the differences become starker, with the RHEA reaching win rates of 7.6 times (for easy games) up to 20 times (for hard games) that of the HPA. For four players the RHEA has at least 27 times the win rate of HPA. It should be noted that at high difficulty neither algorithm performs particularly well: the RHEA wins in 6\% of games with two players, 4\% of games with three players, and 1\% of games with four players. At high difficulty the HPA struggles to find any wins (1.3\% for two players, 0.2\% for three players); therefore, the RHEA outperforms HPA significantly in hard difficulties, and can win some games with four players while the HPA fails consistently.

We should note that in random easy Pandemic games with 4 players the RHEA performs at 19.7\% win rate, while for the same difficulty setting and number of players the same agent reached 40.6\% win rate on the ten testbed setups. The chosen testbeds were admittedly among the easiest that the HPA baseline could solve (as only the $10^3$ initial setups with the highest win rates for HPA were used for clustering). Moreover, as explored in \cite{sfikas2020collaborative} the player order and the order of hidden decks had a strong impact on the behavior of both RHEA and HPA even when the same ten initial game setups were used. 

\section{Discussion}\label{sec:discussion}

The extensive experimentation documented in Section \ref{sec:experiment} demonstrated how a rolling horizon evolutionary algorithm can enhance the performance of the hand-crafted baseline agent for playing \emph{Pandemic}, reaching win ratios as high as 6.3 times those of the baseline in the carefully crafted set of initial game setups. However, such a performance comes at a heavy computational cost; the trade-off analysis has shown that a win rate of 39\% can be reached by agents performing a more realistic 0.5 to 0.6 seconds delay per decision. Moreover, it is evident that performance increases at higher mutation rates which indicates that the mutation operator chosen is not disruptive. This is not surprising, as both the HPA which instantiates (and repairs) the genes and the RPA which applies the mutations are carefully defined and tightly controlled by expert knowledge. That said, the improved performance of RHEA shows that it can modify the expert knowledge baseline substantially and can anticipate better the upcoming challenges through multiple simulations of the forward model.

Most experiments in this paper used the same ten initial setups throughout all the evaluation phases. These initial setups afford a controlled environment on which extensive parameter tuning can take place; during actual simulations the only stochasticity comes from the epidemic cards which cause discarded infection cards to be reshuffled and placed back on the infection deck. While special care was taken to sample representative games across the spectrum (fast games, easy games, etc.), it should be noted that these initial setups were all games that the baseline agent could at least potentially win. 
When testing the RHEA in unknown game setups, varying the number of players and difficulty of the game led to an expected performance drop. However, in harder and less controlled problems the performance of RHEA was even better than the HPA baseline and managed to win some games in cases where the baseline could win none. On the other hand, a RHEA that starts from a more successful initial individual (and uses it for repair) such as the A* algorithm of \cite{sauma2020pandemic} may lead to even better performance and will be explored in future work. However, due to the slight differences between the current testbed and that of \cite{sauma2020pandemic} it is likely that A* is less able to handle the higher branching factors introduced by the operations expert role which adds more research centers on the board that also lead to more traversal options. 

This paper argues that collaborative board games pose a novel challenge for gameplaying AI, however it should be noted that all players in the current Pandemic RHEA are controlled by a single agent. As we note in the introduction, in such games devising a careful plan that is followed to the letter by all players is the norm, which is very close to how the one controller handles different players' turns. That said, it could be interesting to explore multi-agent collaboration, which could also simulate individual goals and priorities. Indicatively, collaborative agents have been tested in digital games \cite{xion2018hogrider,prada2015geometry} as well as in the team-based competitive card game \emph{Hanabi} \cite{canaan2020hanabi}. Such a multi-agent collaboration AI is better suited for games where players have both joint goals and individual goals, such as \emph{Dead of Winter} (Plaid Hat Games, 2014). Another interesting extension concerns human and AI players collaborating in a Pandemic game, similar to \cite{eger2017hanabi}. The main challenge in such a case would be designing an interface for the AI to explain the plan and convince the human players to follow it (and vice versa).

Extensions of this work could re-introduce certain aspects of Pandemic which were omitted for the sake of simplicity, namely the event cards that can be played out-of-turn and three player roles (dispatcher, quarantine specialist, contingency planner). Moreover, RHEAs could be implemented on a less controlled action or state representation, e.g. allowing evolution to choose which cards to share with other players, or keeping previous plans intact (e.g. if another player is going somewhere to share knowledge) when seeding the initial population. While preliminary experiments with multi-objective evolution seemed to yield worse results, it is possible that a larger population rather than a 1+1 evolutionary strategy could result in better performance; this could allow an algorithm such as NSGA-II \cite{Deb2002nsga2} to create a Pareto front of optimistic versus pessimistic tradeoffs. Finally, there is a wealth of collaborative board games which could be tested with this or similar approaches, from simple games with tractable state representations such as \emph{Forbidden Island} (Gamewright, 2010) to games with several layers of stochasticity, inventory management and subsystems such as \emph{Zombicide} (CMON, 2012). As evidenced in this paper, the design patterns of collaborative board games pose many challenges to AI and can lead to breakthroughs which can inform other complex team-based tasks beyond games (e.g.~risk control or scheduling). The recent introduction of the Tabletop Games Framework \cite{gaina2020tabletopgames}, that includes Pandemic, promises that research on gameplaying AI for different types of modern board games will remain strong.

\section{Conclusion}\label{sec:conclusion}

Collaborative board games pose new challenges to artificial intelligence, as agents that handle decision-making in such games must balance between short-term risk mitigation and long-term winning strategies. As evidenced in this paper, a Rolling Horizon Evolutionary Algorithm can perform well in the Pandemic board game when its fitness strikes the right balance between pessimistic (to thwart losing) and optimistic (to reward winning) state evaluations. The highly stochastic nature of Pandemic is also shown to hamper performance when the forward model is not re-initialized and re-evaluated multiple times, as well as when the decision-making horizon is set too far into the future. It should be noted that the evolutionary algorithm operates on a fairly constrained space, which limits its freedom to find novel strategies or control low-level actions such as choosing which cards to discard for moving to remote cities. Future work in collaborative board game play should explore the tradeoffs between a more expressive controller with fewer ad-hoc scripts and the vast and stochastic possibility space that such games invoke.

\section*{Acknowledgements}
We would like to thank Alberto Tonda for early discussions and formalizations of the problem of \emph{Pandemic} agent control. This project has received funding from the European Union’s Horizon 2020 programme under grant agreement No 951911. 

\bibliographystyle{IEEEtran}
\bibliography{tog_pandemic}

\end{document}